\pdfoutput=1
\documentclass[11pt]{article}

\usepackage{naacl2021}

\usepackage{times}
\usepackage{latexsym}

\usepackage[T1]{fontenc}
\usepackage[utf8]{inputenc}

\usepackage{microtype}
\usepackage{balance}

\title{Instructions for NAACL-HLT 2021 Proceedings}

\usepackage{microtype}
\usepackage{multirow}
\usepackage{amsmath}
\usepackage{booktabs}       % professional-quality tables
\usepackage{xargs}
\usepackage{graphicx}
\usepackage[colorinlistoftodos,prependcaption,textsize=small]{todonotes}

\usepackage{multirow}

\usepackage{paralist}

\definecolor{aqua}{rgb}{0.0, 1.0, 1.0}
\definecolor{aquamarine}{rgb}{0.5, 1.0, 0.83}
\definecolor{bittersweet}{rgb}{1.0, 0.44, 0.37}
\definecolor{amber}{rgb}{1.0, 0.49, 0.0}
\definecolor{aureolin}{rgb}{0.99, 0.93, 0.0}

\newcommand{\PredE}{{PredE}}
\newcommand{\EntE}{{EntE}}
\newcommand{\CircE}{{CircE}}
\newcommand{\CorefE}{{CorefE}}
\newcommand{\ConE}{{LinkE}}
\newcommand{\OutE}{{OutE}}
\newcommand{\GramE}{{GramE}}
\newcommand{\NE}{{NE}}
\newcommand{\OthE}{{OthE}}

\newcommand{\Sref}[1]{\S\ref{#1}}
\newcommand{\wrong}[1]{\textcolor{red}{\textbf{#1}}}

\newcommand{\dataset}{FRANK}
\title{Understanding Factuality in Abstractive Summarization with \dataset{}:\\
A Benchmark for Factuality Metrics }

\author{Artidoro Pagnoni  \qquad 
Vidhisha Balachandran \qquad
Yulia Tsvetkov\\
Language Technologies Institute\\
Carnegie Mellon University\\
\texttt{\{apagnoni,vbalacha,ytsvetko\}@cs.cmu.edu}}

\date{}

\begin{document}
\maketitle
\begin{abstract}
%Recent work highlighting the importance of measuring the factuality of automatically generated summaries has motivated a surge of attempts to build metrics to address this need.
Modern summarization models generate highly fluent but often factually unreliable outputs. This motivated a surge of metrics attempting to measure the factuality of automatically generated summaries. 
Due to the lack of common benchmarks, these metrics cannot be compared. Moreover, all these methods treat factuality as a binary concept and fail to provide deeper insights on the kinds of inconsistencies made by different systems. To address these limitations, we devise a \emph{typology of factual errors} and use it to collect human annotations of generated summaries from state-of-the-art summarization systems for the CNN/DM and XSum datasets. Through these annotations we identify the proportion of different categories of factual errors in various summarization models and benchmark factuality metrics, showing their correlation with human judgement as well as their specific strengths and weaknesses.\footnote{Code, data, and online leaderboard will be available at \href{https://github.com/artidoro/frank}{https://github.com/artidoro/frank}}
%We provide insights that will serve future development of both abstractive summarization systems and new factuality evaluation metrics. 

\end{abstract}

\section{Introduction}

%Abstractive text summarization aims at compressing the information of a source document into a rephrased, condensed summary, while ensuring that the source is represented correctly. 
Factuality is defined as a measure of ``whether eventualities are characterized as corresponding to facts, possibilities, or situations that do not hold in the world'' \citep{sauri2008factuality, sauri-event-factuality}. In summarization, this ``world'' is the article, which is taken as ground-truth, and the output summary must be faithful to the article's facts. 
Despite advancements in neural abstractive summarization \citep{xsum, liu-lapata-2019-text, bart}, $\sim$30\% of summaries have factual inconsistencies \citep{fact-aware}. 
With summarization being an integral component of information consumption, this highlights a need for ensuring summarization systems are factually consistent and developing methods for evaluating them.

% \begin{table}[]
%     \centering
%     \small
%     \begin{tabular}{p{\columnwidth}}
%     \toprule
%     \textbf{Short Text}: 
%     Following months of human trials, Pfizer announced its vaccine is more than 90\% successful in preventing Covid-19. It will now proceed to request FDA emergency approval for the vaccine. Other companies developing vaccines are expected to release results of their trials soon.\\
%     % FBI is unlocking a phone in connection with the killing of a couple after it demonstrated it could do so with the San Bernardino case.
%     \midrule
%     \textbf{Summary}: 
%     Pfizer announced its vaccine is at least 90\% successful. \colorbox{amber}{This follows} its request for emergency approval of the drug. \colorbox{aureolin}{The FDA} is expected to release more results soon.\\
%     % \colorbox{amber}{They} said \colorbox{amber}{they} will help the \colorbox{aureolin}{San Bernardino killer} to access IPhones used by \colorbox{aureolin}{the San Bernardino victims}.\\
%     % \midrule
%     % CNN/DM Summary (BERTSum)\\
%     % "BBC claimed H982 FKL was an ``unhappy coincidence''. But maria Cristina Barrionuevo said they were ``provoking people''. She slammed their actions as ``arrogant and disrespectful''. Top Gear producers removed the plate and replaced it with another.\\
%     \bottomrule
%     \end{tabular}
%     \caption{A summary exhibiting two types of factual errors: a \colorbox{amber}{temporal sequence} and an \colorbox{aureolin}{entity} error.}
%     \label{tab:example}
% \end{table}

Common evaluation metrics for summarization based on n-gram overlap -- BLEU, ROUGE, and METEOR \citep{Bleu, rouge, meteor} -- 
are insufficient to measure the factual correctness of summaries and fail to correlate with the human judgements of factuality \citep{falke-etal-2019-ranking, sumcriticaleval}. 
More recent metrics proposed to improve the evaluation of summarization factuality \citep{factcc, feqa, factqa, factentailment} cannot be compared due to the lack of common benchmarks.
More critically, while these approaches differ in the way they model factuality, they all consider factuality as a binary concept, labeling summaries of any length as factual or non-factual. They do not provide any fine-grained understanding of the factual errors made by different systems that could serve as an actionable feedback on a system's limitations.

The binary factuality of a text can be difficult to determine.
\citet{falke-etal-2019-ranking} show relatively low crowd--expert agreement, %is below $\kappa=0.74$, 
indicating the presence of subjectivity in the annotation process. Moreover, not all factual errors are equally important and the number of errors can have a significant impact on the perceived factuality of a text. This suggests that non-factuality should be modeled as a multi-dimensional construct and not a label. 
%This suggests that non-factuality is a nuanced concept.

\begin{figure*}
    \centering
    \includegraphics[width=\linewidth]{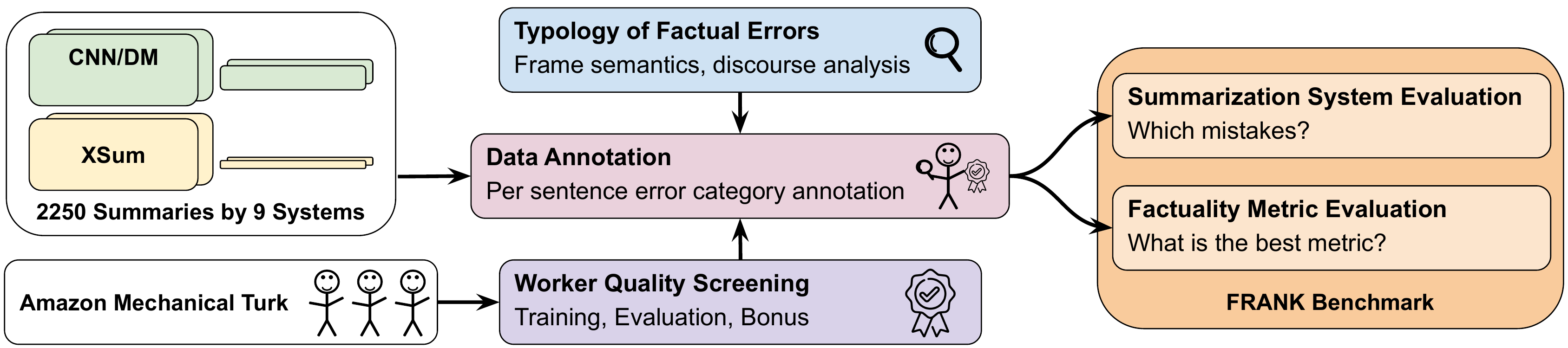}
    \caption{We propose a linguistically grounded typology of factual errors. We select crowd workers to annotate summaries from two datasets according to this typology achieving near perfect agreement with experts. We collect \dataset{}, the resulting dataset, to benchmark factuality metrics and state-of-art summarization systems. }
    \label{fig:methodology}
\end{figure*}

In this work, we propose a linguistically motivated typology of factual errors for fine-grained analysis of factuality in summarization systems (\Sref{sec:typology}). Our typology is theoretically grounded in frame semantics \citep{fillmore1976frame, palmer2005proposition} and linguistic discourse theory \citep{brown1983discourse}. 
%We propose seven categories of possible factual errors which quantify factuality in a nuanced way. 
It provides several benefits. First, we find that decomposing the concept of factuality in (relatively) well-defined and grounded categories makes the final binary decision more objective leading to near perfect agreement between crowd and expert annotators ($\kappa = 0.86$). Second, this approach provides some measure of the degree of non-factuality both in terms of the quantity and the category of factual violations that appear in the text. This typology also provides us with the means to categorize the types of errors made by summarization systems, helping us gain deeper insights than simply categorizing content as factual or hallucinated.

%We use this definition of facts and factuality to assemble a linguistically grounded typology of factual errors in summarization (\Sref{sec:typology}). 
We define an annotation protocol of factuality based on our typology and collect a dataset of human judgements over a diverse set of model generated summaries on the CNN/DM \citep{cnn-dm} and XSum \citep{xsum} datasets (\Sref{sec:humanevalstudy}). 
Through this dataset, we aim to both assess the factuality of summarization systems and benchmark recently proposed factuality metrics. In \Sref{sec:model_analysis} we discuss various state-of-art models and show a detailed analysis of the factual errors they make. Finally, in \Sref{sec:metric_evaluation} we evaluate multiple summarization metrics against our benchmark and show their strengths and weaknesses in detecting specific types of factual errors. \autoref{fig:methodology} shows an overview of this work.

\begin{table*}[ht]
    \small
    \begin{tabular}{  l  p{2.5cm} p{5.2cm}  p{5.5cm} }
        \toprule
& \textbf{Category}
& \textbf{Description}
& \textbf{Example} \\\midrule
\textbf{\PredE{}}
    & Relation Error
    & The predicate in the summary statement is inconsistent with the source article. 
    & \textit{The Ebola vaccine \wrong{was rejected} by the FDA in 2019.} 
    \\\hline
\textbf{\EntE{} }
    & Entity Error          
    & The primary arguments (or their attributes) of the predicate are wrong.
    & \textit{The \wrong{COVID-19 vaccine} was approved by the FDA in 2019.} 
    \\\hline
\textbf{\CircE{} }      
    & Circumstance Error 
    & The additional information (like location or time) specifying the circumstance around a predicate is wrong. 
    & \textit{ The first vaccine for Ebola was approved by the FDA in \wrong{2014}.}
    \\\hline
\textbf{\CorefE{}} 
    & Coreference Error
    & A pronoun/reference with wrong or non-existing antecedent.
    & \textit{The first vaccine for Ebola was approved in 2019. \wrong{They} say a vaccine for COVID-19 is unlikely to be ready this year.}
    \\\hline
\textbf{\ConE{}} 
    & Discourse Link Error
    & Error in how multiple statements are linked together in the discourse (for example temporal ordering/causal link).
    & \textit{To produce the vaccine, scientists have to show successful human trials, \wrong{then} sequence the DNA of the virus.}
    \\\hline
\textbf{\OutE{}}    
    &  Out of Article Error
    &  The statement contains information not present in the source article.
    &  \textit{\wrong{China} has already started clinical trials of the COVID-19 vaccine.}
    \\\hline
\textbf{\GramE{}} 
    & Grammatical Error     
    & The grammar of the sentence is so wrong that it becomes meaningless.
    & \textit{The Ebola vaccine \wrong{accepted have already started.}}
    \\\bottomrule
    \end{tabular}
    \caption{Typology of factual errors. Original text for the examples: 
    \textit{The first vaccine for Ebola was approved by the FDA in 2019 in the US, five years after the initial outbreak in 2014. To produce the vaccine, scientists had to sequence the DNA of Ebola, then identify possible vaccines, and finally show successful clinical trials. Scientists say a vaccine for COVID-19 is unlikely to be ready this year, although clinical trials have already started. 
    %Over 200,000 people have died of COVID-19 in the US.
    }}
    \label{tab:framework}
\end{table*}

\section{Typology of Factual Errors}
\label{sec:typology}
Previous studies of factuality in summarization only distinguish factual and hallucinated content \citep{sumcriticaleval, factentailment} and provide limited insights on the fine-grained types of factual errors. In the simplest case, factual errors appear within a single proposition. However, as summaries include several sentences, discourse markers describe relations across propositions. These cross-sentence links, such as causality or temporal ordering, can introduce inconsistencies with the article. Furthermore, information in the summary should be verifiable given the article. 
This understanding outlines different levels of linguistic structure where factual mistakes can arise in summaries: at the semantic frame level, at the discourse level, or because the content cannot be verified.
Below we define a typology of factual errors further detailing these three levels. This typology is theoretically grounded in frame semantics \citep{fillmore1976frame, baker-etal-1998-berkeley, palmer2005proposition} and linguistic discourse analysis \citep{brown1983discourse}. Examples for each category are shown in \autoref{tab:framework}.

\subsection{Semantic Frame Errors}
% Single fact errors.
A \emph{semantic frame} is a schematic representation of an event, relation, or state, which consists of a predicate and a list of participants, called frame elements \cite{baker-etal-1998-berkeley}. A semantic frame has both core and non-core frame elements (FE). Core frame elements are essential to the meaning of the frame, while non-core (e.g. location, time) provide additional descriptive information.
Our first three categories capture factual errors in each of these components (frame, core and non-core FE) respectively.

% A \textit{frame} is defined a relation based on a predicate that describes the functional association between entities based on Frame Semantics \citep{fillmore1976frame}.
% This is akin to the notion of frame in Frame Semantics \citep{fillmore1976frame}. 
% A predicate can be viewed as a function that takes one or more entities as arguments. The first three classes of errors involve a single frame following PropBank arguments closely.
%F1

\paragraph{Predicate Error (\PredE{}):} Category \emph{\PredE{}} encompasses errors where the predicate in a summary statement is inconsistent with the source text. More generally, this represents cases where the frame from a summary statement does not align with what is expressed in the source text. %This includes negation errors.
% implies an inconsistent functional relation. Note that different from PropBank, we consider negations as integral specification of the predicate, so negation errors would be part of this class.
%F2 
\paragraph{Entity Error (\EntE{}):} Category \emph{\EntE{}} captures errors where the primary arguments (like entities) of the predicate are wrong or have the wrong attributes, although the relation was expressed in the original text. More generally, these account for cases where the core frame elements in a frame are wrong. This also captures directionality errors where the elements are interchanged (similar to agent-patient swap).

% In PropBank terms, the verb-specific numbered roles (including Arg0 and Arg1 or agent and patient) are incorrect. This category captures directionality errors such as agent-patient swap, where the entities and their functional relation are independently correct, but the entities are assigned to the wrong arguments of the relation.\todo{Maybe expand on what we mean by wrong attributes: quantifiers, or qualifying expressions that specify the entity.}
%F3
\paragraph{Circumstance Error (\CircE{}):} In additional to the core arguments, predicates can be further specified using additional information or attributes that describe the circumstance in which the arguments and predicates interact (e.g. location, time, manner, direction, modality). Category \emph{\CircE{}} captures errors where one or more such attributes (non-core frame elements within a frame) are wrong.

% general, adjunct-like arguments (in PropBank this falls under the modifier tag ArgM). Such arguments specify the location, time, manner, direction, modality, etc. of the predicate. We use \emph{\CircE} to capture such errors.

\subsection{Discourse Errors}
% Discourse errors
%F4
The communicative intent of an author is also expressed through relations that hold between parts of the text. Factual errors in summarized text can often extend beyond a single semantic frame introducing erroneous links between discourse segments. 
Below we outline such categories of errors which are grounded in
discourse analysis and rhetorical structure theory (RST) \citep{brown1983discourse,mann1988rhetorical}. RST is an elaborate system for annotating coherence relations in discourse. Some examples of such relations include: ``Elaboration'', ``Background'', ``Motivation'', and ``Volitional Cause''. Here we depart from semantic frame terminology as its rooting in a single frame does not allow us to represent such errors.

\paragraph{Coreference Error (\CorefE):} Category \emph{\CorefE} accounts for errors where pronouns and other types of references to previously mentioned entities either are incorrect or have no clear antecedents, making them ambiguous. 

\paragraph{Discourse Link Error (\ConE):} Category \emph{\ConE} encompasses errors involving a discourse link between different statements. These include errors of incorrect temporal ordering or incorrect discourse links (e.g. RST relations, discourse connectors) between statements. 
% Examples of such errors include wrong causal relation or temporal order between facts.

\subsection{Content Verifiability Errors}
% \todo{Rename: Can we think of a better title? EDIT: changed `Basic Content Errors` to `Verifiability Errors` Is this better? Alternatively `Form errors`}
% Facts should be be verifiable given the article: the information should be contained in the article, and they should be well formed. 
Often statements in a summary cannot be verified against the source text due to difficulty in aligning them to the source. Below we outline two categories of errors for such cases.

% Other
\paragraph{Out of Article Error (\OutE):}
Since summaries of a document should only contain information that can be deduced from the original text, we include a category for such errors \emph{\OutE} (prior work refers to this as extrinsic hallucinations \citep{factentailment}).

\paragraph{Grammatical Error (\GramE):}
We use \emph{\GramE} to categorize statements that are not well formed. When grammatical mistakes make the meaning of a statement incomprehensible or ambiguous, it cannot be verified against the source and is thus considered trivially wrong. Minor grammatical errors are acceptable.

% We show empirically that these categories have the desirable characteristics of being complete and mutually exclusive. \todo{Add reference to section.}

Finally, for completeness in our annotation exercise, we add two additional categories \textbf{Others (\OthE{})} for factually errors that do not correspond to any of the above categories and \textbf{Not an Error (\NE{})} for statements that do not contain any errors.

\section{Dataset Creation}
\label{sec:humanevalstudy}
Beyond theoretical grounding, we empirically verify our typology through large scale human annotations of five abstractive summarization models on the CNN/DM dataset and four on the XSum dataset. Through our dataset, we aim to have a broad coverage of different types of errors made by neural summarization systems, with human judgements on their fine-grained factuality errors.

\paragraph{Annotation Data}
For the annotation, we include model summaries from CNN/DM and XSum datasets as they present different characteristics.
% CNN/DM articles are twice as long as XSum articles.
CNN/DM summaries are longer, with three sentences on average, while XSum has only single sentence summaries. Having longer summaries is crucial to identify discourse level errors. On the other hand, XSum summaries are more abstractive and include more factual errors on average \citep{factentailment}. For a diverse set of model summaries, we collect publicly available model outputs from different summarization models with differing factuality capabilities. For the CNN/DM dataset, we use model outputs from a LSTM Seq-to-Seq model (S2S) \citep{rush}, a Pointer-Generator Network (PGN) model \citep{pgn}, a Bottom-Up Summarization (BUS) model \citep{bus}, a Bert based Extractive-Abstractive model (BertSum) \citep{liu-lapata-2019-text} and a jointly pretrained transformer based encoder-decoder model BART \citep{bart}. For the XSum dataset, we collect model outputs from a Topic-Aware CNN Model \citep{xsum}, a Pointer-Generator Network (PGN) model, a randomly initialized (TransS2S) \citep{vaswani2017attention} and one initialized with Bert-Base (BertS2S) \citep{bert}.\footnote{As we use publicly available model outputs, the summaries across different datasets are from different models owing to their availability.} Details of the models used are provided in \Sref{sec:model-details}.

% For a diverse set of model summaries, we collect model outputs from three popular families of summarization models. From the LSTM family, we chose a LSTM Seq-to-Seq model (S2S) \cite{}, a Pointer-Generator Network (PGN) model \cite{}, a Bottom-Up Summarization (BUS) model \cite{}. From the Convolutional model family we choose the Topic-Aware CNN Model \cite{}. From the Transformer family, we chose two Transformer Seq-to-Seq models one randomly initialized (TransS2S) \cite{} and one initialized with Bert-Base (BertS2S) \cite{}, a Bert based Extractive-Abstractive model (BertSum) \cite{} and a jointly pretrained transformer based encoder-decoder model BART \cite{}. Details of the models used are provided in \Sref{sec:model-details}. We choose these models because their outputs have varying degrees of abstraction.
% , e.g., PGN is less abstractive, as it tends to copy longer  sequences from input text \citep{bus, balachandran2020structsum}. We include the models that were initialized on large-scale language modeling to evaluate whether it facilitates  higher factuality. 
% For each dataset, we chose publicly available model outputs: S2S, PGN, BUS, BERTSum, and BART for the CNN/DM dataset, and PGN, TConvS2S, TransS2S, and BERTS2S for the XSum dataset. 

\begin{figure*}[ht!]
    \centering
    \includegraphics[width=\linewidth]{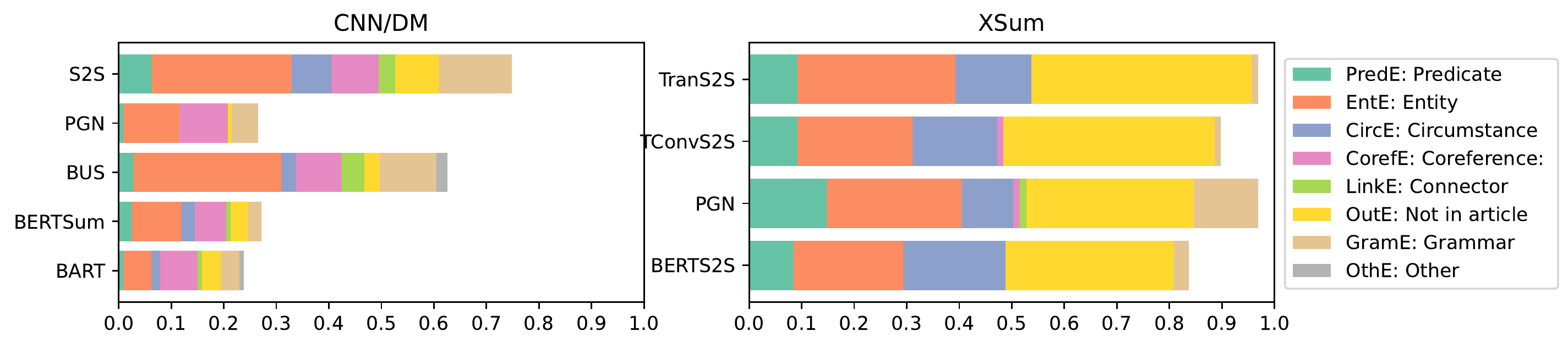}
    \caption{Proportion of summaries with factual errors based on collected annotations, with breakdown of the categories of errors within. Full specification of categories of errors in \autoref{tab:framework}.}
    \label{fig:factcategory}
\end{figure*}

\paragraph{Annotation Collection}
% \citet{falke-etal-2019-ranking} have shown in their experiments that three or more crowd source annotators have substantial agreement with expert annotators of factuality, (with Cohen's $\kappa>0.65$). We therefore choose to use three annotators for each data point.
Using the above model generated summaries, we collect human annotations from three independent annotators
% \footnote{\citet{falke-etal-2019-ranking} have shown that three or more crowd source annotators have strong agreement with expert annotators of factuality, (with Cohen's $\kappa>0.65$).} 
for 250 articles from each dataset (with a total of 1250 model outputs on CNN/DM and 1000 on XSum). 
We annotate each sentence of a summary to break the judgement of factuality into smaller units. We present sentences in the context of the entire summary to identify discourse errors spanning multiple sentences. Annotations are a two step process: for each sentence in the summary, the annotator first selects whether the sentence is factual, and if marked not factual, identifies the category of each error based on our typology.
%We annotate each sentence of a summary individually as previously recommended by \citet{feqa}. 
%However, we depart from prior work by presenting the sentence in the context of the entire summary. 
\footnote{We experimented with Likert scale evaluation of full summaries in a pilot study. Such an annotation would not provide precise information about where in the summary an error appears and also resulted in lower agreement. Hence, we opted for sentence level judgements.} A sentence can be annotated with more than one category of errors to account for multiple errors within a sentence. We conduct the annotation task on the Amazon Mechanical Turk (MTurk) platform. To achieve high quality crowd-sourced annotations, we build an intuitive interface\footnote{We make the interface available for future human annotations that follow our typology} which combines:
\begin{compactenum}
    \item \textbf{Clear Instructions: }We explain the annotation scheme without assuming linguistic knowledge and give several examples for each category.
    \item \textbf{Training and Evaluation: }We setup training tutorials for first time users to train and provide feedback on the task. We also setup a qualification test which tests their understanding of our annotation scheme and require annotators to obtain >85\% score to qualify. Further, we continuously evaluate annotators during the task against artificially generated factual errors to ensure continued high quality.
    \item \textbf{Fair Pay and Bonus: } All workers are paid 50\% more than the average American minimum wage. We offer bonuses for scores of 60\% or above on the continuous evaluation, and for completing sets of 10 annotations.
    % \footnote{We observe that bonuses increase the percentage of users with high continuous evaluation scores ($<$10\% blocked users with bonuses versus 30\% without bonuses).}
\end{compactenum}

Further details on our interface are added in \Sref{sec:annotation_details}

\paragraph{Inter-Annotator Agreement: }
We report inter-annotator agreement in terms of Fleiss Kappa $\kappa$ \citep{fleiss1971measuring}. %, which accounts for several annotators.
Following \citet{feqa}, we report the percentage $p$ of annotators that agree with the majority class. 
Each datapoint in our dataset corresponds to a sentence in a summary. We compute agreement on all 4942 annotated sentences.
On the annotation of whether a sentence is factual or not we obtain $\kappa=0.58$, with $p=91\%$ of annotators agreeing with the majority class. As a comparison, \citet{feqa} reports $p=76.7\%$ average agreement.
When all three annotators agree that a sentence is not factual, we obtain $\kappa=0.39$ with $p=73.9\%$ of annotators agreeing with the majority class on the eight category annotation (seven categories of errors and ``other'') which indicate a moderate agreement. 
% These numbers indicate a moderate agreement potentially due to the large number of classes and crowd sourced annotations from non-domain experts. 
% We do not discard any data points that have low agreement to provide a fair set up that includes difficult examples. 
% Other collections of human judgements of factuality do not mention the number of datapoints that were filtered due to low agreement.

\paragraph{Agreement with Domain Expert: }
We measure agreement between the majority class of the three annotators and one expert annotator on 201 datapoints (10 summaries from CNN/DM and 10 summaries from XSum). We find a Cohen Kappa of $\kappa = 0.86$ indicating nearly perfect agreement. Previous work found agreement of $\kappa = 0.65$ between three crowd annotators and expert annotations of factuality \citep{falke-etal-2019-ranking}. Even with more than nine workers, they report agreement with expert annotations of at most $\kappa = 0.74$. This improvement validates the robustness of our annotation interface and protocol which achieves higher agreement with fewer workers.  

\section{Summarization Model Analysis}
\label{sec:model_analysis}
We evaluate the performance of different summarization models in terms of factuality. \autoref{fig:factcategory} visualizes the percentage of summaries with factual errors for each category model and dataset, with a breakdown of proportion of different error types within each.  A summary is considered incorrect if it contains at least one sentence with a factual error. A sentence contains a factual error if the majority of annotators indicate the presence of an error (here we do not consider annotations where all three annotators disagree on the category). 
% We observe that there is a different distribution of errors for different models, as well as for different datasets.

% \paragraph{How factual are generated summaries across different datasets? }First, we observe that 63.1\% of the summaries annotated contain a factual error. 
% % On the CNN/DM dataset, even state-of-the-art pretrained models like BertSum and Bart have x\% and y\% summaries containing factual errors respectiveky.  
% Even state-of-the-art models (in terms of ROUGE) on the CNN/DM dataset still have around 30\% of summaries containing factual errors. 
% The XSum dataset poses more significant challenges in terms of factuality as even the best model produces factual errors in 80\% of the summaries. For the CNN/DM dataset, the most frequent class of errors is \textbf{\EntE{}} (wrong entity) next is \textbf{\CorefE{}} (wrong coreference) and for the XSum dataset it is \textbf{\OutE{}} (information not in article). Most notably, there are no discourse errors (\textbf{\CorefE{}, \ConE{}}) in the XSum dataset because the data only contains single sentence summaries.

\paragraph{How factual are generated summaries across different datasets? }From our annotations, we observe that 60\% of the summaries that were annotated contain at least one factual error. From \autoref{fig:factcategory}, we see that the XSum dataset has more factually incorrect model summaries (92\%) than CNN/DM (43\%). It poses more significant challenges in terms of factuality as all models produce > 80\% summaries with factual errors, with the best model (BertS2S) producing 83\% wrong summaries. On the CNN/DM dataset, while state-of-the-art pretrained models like BERTSum and BART have better factuality numbers, the percentage of factually incorrect summaries is still high (23\% for BERTSum and 27\% for BART).
% Even state-of-the-art models (in terms of ROUGE) on the CNN/DM dataset still have around 30\% of summaries containing factual errors. 
The proportion of errors across different categories vary widely between the two datasets. For the CNN/DM dataset, the most frequent classes of errors are Entity Error (\EntE{}) and Coreference Error (\CorefE{}). For the XSum dataset they are Out of Article Error (\OutE{}) and Entity Error (\EntE{}). Note that there are no discourse errors (\CorefE{}, \ConE{}) in the XSum dataset because the data only contains single sentence summaries. Additionally, we observe that \OthE{} makes up a very small percentage ($\sim$ 1\%) of errors overall showing that our typology is \emph{complete} with most errors being mapped to one of our existing categories.

% This supports our decision to include different classes of models and both CNN/DM and XSum datasets. This produces a more diverse set of examples and avoids biasing the benchmark towards a single model class or dataset. 
% \paragraph{How factual are generated summaries across different models? } We observe that different models have different distributions of errors. PGN, BERTSum, and BART have relatively more coreference errors than BUS and Seq2Seq on the CNN/DM dataset. We observe that models pre-trained on language modeling tasks tend to perform better, as also noted by \citet{feqa}. On the CNN/DM dataset BERTSum and BART display half the error rate than BUS. However, \autoref{tab:factcategory} shows that, for these two models, the better error-rate on the CNN/DM dataset come from modeling improvements at the frame level (\textbf{\PredE{}, \EntE{}, \CircE{}}) while modeling the discourse remains a challenge. Errors (\textbf{\CorefE{}, \ConE{}}) now account for a higher proportion of errors in BERTSum and BART compared to the other models. We also observe that PGN has fewer than half the error rate of BUS while both models share the same core architecture. One important distinction between these models is that PGN is more extractive, and has been shown to copy large portions of text \citep{bus, balachandran2020structsum}. Intuitively, extractive models have fewer chances of introducing errors in the summaries.

\paragraph{How factual are generated summaries across different models? } From \autoref{fig:factcategory}, we observe that LSTM based models like S2S and BUS generate many incorrect summaries. Interestingly, PGN on CNN/DM has fewer summaries with factual errors (26\%) compared to  S2S (74\%) and BUS (62\%) potentially due to the extractive nature of CNN/DM and the copy based objective in PGN. PGN has been previously shown to produce highly extractive summaries on CNN/DM copying large portions of text (often entire sentences) \citep{bus, balachandran2020structsum}. On the more abstractive dataset XSum, PGN produces $> 96\%$ factually incorrect summaries. We also observe that large-scale pretrained models improve factuality on both datasets, as also noted by \citet{feqa}, with more significant gains on CNN/DM. On CNN/DM, BERTSum and BART display half the error rate of BUS. In contrast, on XSum, BertS2S improves over non-pretrained models by $\sim$ 10\% only, showing that XSum poses a significant challenge for factuality even in pretrained models. 

Different models also exhibit different distributions in the error categories. LSTM based models have higher proportion of Grammatical Errors (\GramE{}) while transformer and CNN based models have a lower proportion. 
For pretrained transformer models, we observe that the improved error-rate on the CNN/DM dataset can be attributed to improvements at the frame level (\PredE{}, \EntE{}, \CircE{}) while the discourse level errors still remain a challenge. Errors \CorefE{}, \ConE{} account for a higher proportion of errors in BERTSum and BART compared to the other models. 
%Similarly on the XSum dataset, improved factuality for BERTSum can be attributed to fewer Predicate Error (\PredE{}), Entity Error (\EntE{}) and Grammatical Error (\GramE{}), while Circumstance Error (\CircE{}) and Out of Article Error (\OutE{}) are still high owing to the highly abstractive nature of the dataset.
% PGN, BERTSum, and BART have relatively more coreference errors than BUS and Seq2Seq on the CNN/DM dataset. 

\begin{table*}
\centering
\resizebox{\linewidth}{!}{
\begin{tabular}{l| c c c c | c c c c | c c c c }
\toprule
 & \multicolumn{4}{c|}{All data} & \multicolumn{4}{c|}{CNN/DM} & \multicolumn{4}{c}{XSum} \\
\midrule
\multirow{2}{*}{Metrics} & \multicolumn{2}{c}{Pearson} & \multicolumn{2}{c|}{Spearman} & \multicolumn{2}{c}{Pearson} & \multicolumn{2}{c|}{Spearman} & \multicolumn{2}{c}{Pearson} & \multicolumn{2}{c}{Spearman} \\
&$\rho$ & p-val & $r$ & p-val &$\rho$ & p-val & $r$ & p-val &$\rho$ & p-val & $r$ & p-val  \\
\midrule
BLEU & 0.10 & 0.00 & 0.07 & 0.00        & 0.08 & 0.01 & 0.08 & 0.01        & 0.14 & 0.00 & 0.20 & 0.00\\
METEOR & 0.14 & 0.00 & 0.11 & 0.00      & 0.12 & 0.00 & 0.10 & 0.00      & 0.15 & 0.00 & 0.10 & 0.00\\
Rouge-1 & 0.14 & 0.00 & 0.10 & 0.00     & 0.12 & 0.00 & 0.10 & 0.00     & 0.15 & 0.00 & 0.09 & 0.01\\
Rouge-2 & 0.12 & 0.00 & 0.08 & 0.00     & 0.08 & 0.00 & 0.07 & 0.01     & 0.17 & 0.00 & 0.14 & 0.00\\
Rouge-L & 0.13 & 0.00 & 0.09 & 0.00     & 0.11 & 0.00 & 0.09 & 0.00     & 0.16 & 0.00 & 0.10 & 0.00\\
\midrule
OpenIE & 0.11 & 0.00 & 0.02 & 0.36      & 0.16 & 0.00 & 0.15 & 0.00      & 0.00 & 0.93 & -0.45 & 0.00\\
BERTS P & \textbf{0.27} & 0.00 & 0.24 & 0.00     & 0.35 & 0.00 & 0.29 & 0.00     & \textbf{0.18} & 0.00 & 0.09 & 0.00\\
BERTS R & 0.14 & 0.00 & 0.13 & 0.00     & 0.21 & 0.00 & 0.17 & 0.00     & 0.07 & 0.03 & 0.03 & 0.38\\
BERTS F1 & 0.24 & 0.00 & 0.21 & 0.00    & 0.32 & 0.00 & 0.26 & 0.00     & 0.15 & 0.00 & 0.06 & 0.05\\
FEQA & 0.00 & 0.83 & 0.01 & 0.60        & -0.01 & 0.76 & -0.01 & 0.72        & 0.02 & 0.45 & 0.07 & 0.04\\
QAGS & 0.06 & 0.00 & 0.08 & 0.00        & 0.13 & 0.00 & 0.09 & 0.00        & -0.02 & 0.48 & 0.01 & 0.65\\
DAE & 0.16 & 0.00 & 0.14 & 0.00     & 0.25 & 0.00 & 0.24 & 0.00     & 0.04 & 0.16 &\textbf{ 0.28} & 0.00\\
FactCC & 0.20 & 0.00 &\textbf{ 0.30} & 0.00      & \textbf{0.36} & 0.00 & \textbf{0.33} & 0.00      & 0.07 & 0.02 & 0.25 & 0.00\\
\bottomrule
\end{tabular}
}
\caption{Partial Pearson correlation and Spearman rank correlation  coefficients  and  p-values  between  human judgements and metrics scores. Comparisons should be made along with the pairwise Williams test found in \autoref{tab:metric-metric-results}.}
\label{tab:results}
\end{table*}

\section{Factuality Metric Evaluation}
\label{sec:metric_evaluation}
We propose the \dataset{} dataset resulting from the human annotation study as a common benchmark to assess different factuality metrics. We provide an evaluation protocol of factuality metrics, which controls for dataset biases, and a fine grained analysis of the strengths of each metric.

\subsection{Benchmark}
The \dataset{} benchmark provides a diverse dataset for evaluating various metrics on their ability to capture factual errors. Notably, our benchmark has \emph{factual error diversity}, as it covers all types of errors described in the typology in \Sref{sec:typology}, and \emph{data diversity} as it combines 2250 summaries from different systems and datasets. Our annotations go beyond binary labels of factuality on a summary by providing fine-grained category annotations for every sentence. This allows us to determine how well each metric can capture each type of error. Furthermore, through averaging of sentence level judgements, we can also obtain a factuality scores (0 to 1 range) for a summary. To measure the degree that automated metrics capture a certain characteristic, we compute their correlation with human judgements and report Pearson correlation and Spearman rank correlation along with their p-values. 

We evaluate different classes of metrics against the \dataset{} benchmark. We select four general summarization metrics. ROUGE, BLEU, and Meteor are n-gram based metrics and computed with respect to the reference summary. BERTScore \citep{bertscore} computes BERT \citep{bert} contextual embeddings on summary and source article and measures distances between matched embeddings. 
We select five metrics focused on factuality. As \citet{goodrich19}, we use a simple OpenIE \citep{openie} baseline. This involves extracting OpenIE triples and matching them through sentence embeddings \citep{sentence-bert}. FactCC \citep{factcc} and DAE \citep{goyal2020evaluating} are entailment based metrics. FactCC operates with sentences as claims, while DAE uses dependency level entailment. FEQA \citep{feqa} and QAGS \citep{factqa} are two question answering and generation metrics (QGA). More details on the differences between these metrics is in \Sref{sec:metrics_appendix}.
\begin{figure}
    \centering
    \includegraphics[width=\columnwidth]{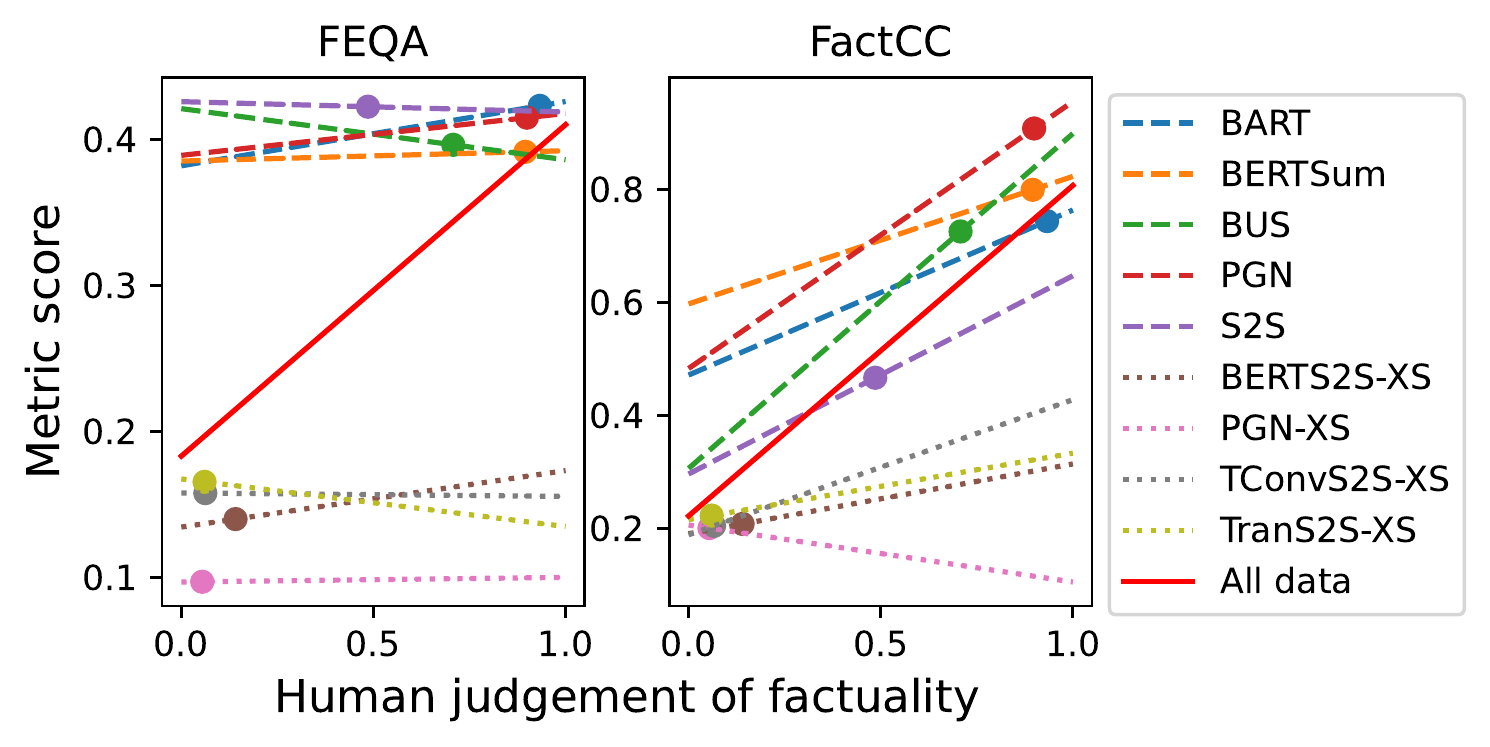}
    \caption{Correlation between metrics and human judgement on subsets of  data. The $x$ and $y$ axis represent the human judgement the metric scores respectively. The red line is a linear regression fitted on full data. Each dotted line is a linear regression fitted on a model-dataset subset. Each colored point has coordinates equal to average factuality judgement, and metric score for its corresponding partition.}
    \label{fig:partial_correlations}
\vspace{-3mm}
\end{figure}
% \subsection{Correlation with Human Judgements}

% In order to have a diverse benchmark, we have included outputs from different systems and datasets. We show that this high diversity brings dataset biases.
\subsection{Controlling for Dataset Biases} Since our benchmark contains diverse summaries from different datasets and models, dataset biases can hamper accurate reporting. 
% A simple test to reveal dataset biases is to verify that properties that show on the overall dataset are also valid in subsets of the dataset. 
% Below, we show how correlations computed on the overall dataset can be misleading due to biases and propose the usage of partial correlations to account for this.
% \paragraph{Biases in Factuality Datasets} 
In \autoref{fig:partial_correlations}, we visually show correlations between two factuality metrics (FEQA and FactCC) and human judgement on the entire data and on partitions of the data. 
% The red line is the result of fitting a linear regression on the entire data. Each dotted line corresponds to fitting a linear regressions on the subset of summaries, where each subset is produced by a different summarization system on a given dataset. In both cases, the $x$ axis represents the human judgement of factuality of summaries, and the $y$ axis represents the score outputted by the metric being evaluated. 
For both metrics, we notice that the slope (an unscaled measure of correlation) of the line fitted through the entire data (red line) is significantly larger. In FEQA, the dotted lines (fitted on subsets of the data of each model and dataset) are almost horizontal. This likely indicates the presence of a confounding variable associated with the properties of each system and dataset. This can lead to false measures of high correlation if not accounted for. To address this, we suggest to control for confounding variables using partial correlations. We include details on partial correlations in the Appendix. In this case, both the system and the dataset are taken to be confounding variables.

\subsection{Results} 
In \autoref{tab:results}, we report the partial Pearson correlation and Spearman rank correlation coefficients with human judgements for each metric, along with their $p$-values indicating statistical significance.

\paragraph{How do different metrics correlate with human judgements? } From \autoref{tab:results} we observe that all metrics exhibit low correlations with human judgements of factuality. The best metrics overall are FactCC with 0.20 Pearson and 0.30 Spearman correlation and BERTScore P with 0.27 Pearson and 0.35 Spearman correlation. Interestingly, we observe that general summarization metrics BLEU, Rouge, and METEOR, and the OpenIE baseline have statistically significant correlations with factuality, close to FactCC ($\rho=0.14$ for Rouge-1 and METEOR versus $\rho=0.20$ for FactCC). The entailment metrics (FactCC and DAE) and contextual embedding method (BERTScore) have the highest correlations and are statistically significant. The two QGA metrics have lower overall correlation. FEQA's correlation is not statistically significant. QAGS has low, but significant correlation of $\rho=0.06$. 

\begin{figure*}
    \centering
    \includegraphics[width=\linewidth]{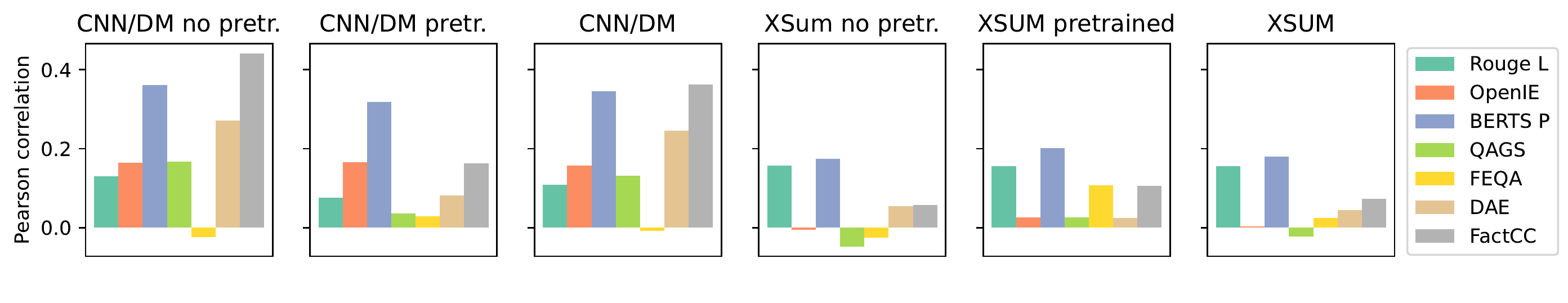}
    \caption{Partial Pearson correlation on different partitions of the data. Entailment metrics have highest correlation on pretrained models in the CNN/DM dataset. Their performance degrades significantly on XSum.}
    \label{fig:correlation_model_data_analysis}
\end{figure*}

\paragraph{How well do different metrics capture errors in different datasets?}
In \autoref{fig:correlation_model_data_analysis}, we observe that entailment metrics have significantly higher partial Pearson correlation on the CNN/DM dataset than XSum where their correlation is reduced by a factor of four. QAGS and the OpenIE baseline have similar behavior. This suggests that these metrics capture the error types from CNN/DM better that those from XSum. Specifically, XSum has uniquely high Out of Article (\OutE{}) errors which they might not capture well. This also highlights the importance of data diversity in building and benchmarking factuality metrics to avoid overfitting to certain types of errors.

\paragraph{How well do different metrics capture errors from pretrained and non-pretrained models?}
On the CNN/DM dataset we observe that entailment metrics and QAGS perform significantly better on non-pretrained models. This indicates that the artificial factual errors on which entailment metrics are trained on are closest to the mistakes that non-pretrained models make. This also suggests that the errors made by pretrained models might be more difficult to capture by these metrics. These trends are less clear on the XSum dataset which we again attribute to high Out of Article (\OutE{}) errors in the pretrained and non-pretrained models (ref \autoref{fig:factcategory})

% \paragraph{Hotelling-Williams test}
% The correlation numbers should be read in combination with the pairwise Hotelling-Williams test result in \autoref{tab:williams} in the appendix. The highlighted numbers indicate pairs of models for which the difference in correlation is statistically significant. 

% \begin{figure}
%     \centering
%     \includegraphics[width=\columnwidth]{plots/partial_correlation_plot_model_name.pdf}
%     \caption{Correlation between metrics and human judgement on the entire data and on partitions of the data. The $x$ axis represents the human judgement of factuality of summaries, and the $y$ axis represents the metric scores. The red line is the result of fitting a linear regression on the entire data. Each dotted line corresponds to fitting a linear regressions on the subset of summaries of each model-dataset setting. Each colored point has coordinates equal to average factuality judgement, and metric score for its corresponding partition. Correlation on all data (steep red line) is high, but on partitions of the data it is lower (dotted lines not as steep) revealing dataset biases leading to high correlation. In the legend, XS indicates XSum dataset.}
%     \label{fig:partial_correlations}
% \end{figure}
\begin{figure}
    \centering
    \includegraphics[width=\columnwidth]{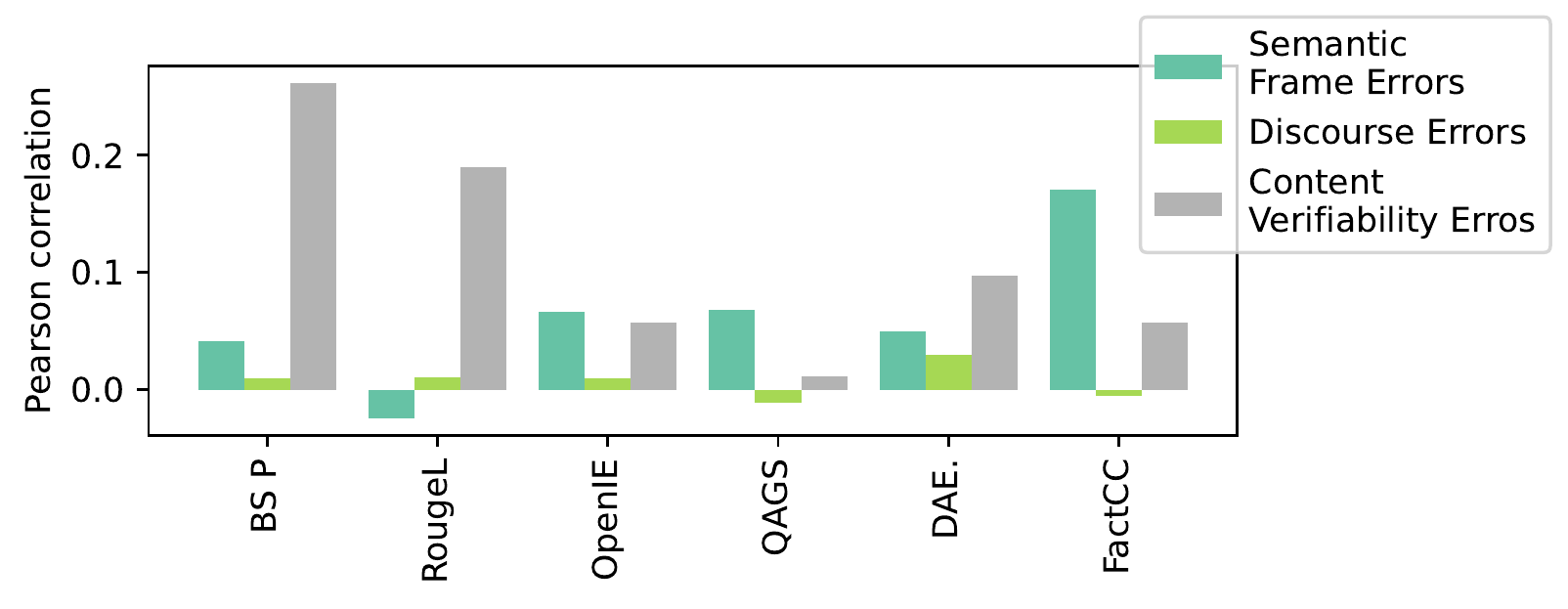}
    \caption{Variation in partial Pearson correlation when omitting error types. Higher variation indicates greater influence of an error type in the overall correlation.}
    \label{fig:correlation_category_analysis}
\vspace{-3mm}
\end{figure}

\subsection{Error Analysis}
% The \dataset{} dataset contains annotations for different types of factual errors (full typology in \Sref{sec:typology}). 
\autoref{fig:correlation_model_data_analysis} shows partial Pearson correlation on six subsets of the data. To understand capabilities of metrics across the broad categories of errors (semantic frame errors, discourse errors, and content verifiability errors) we perform an ablation study. For each category, we compute the variation in partial correlation with errors from that category omitted. In \autoref{fig:correlation_category_analysis}, we visualize the influence of a given type of error using the variation for each metric and category. A higher positive bar indicates that the error type was a significant contributer to the overall correlation (or metric highly correlates with error) causing the correlation without it to drop.

\paragraph{General Summarization metrics}
Unsurprisingly, we observe that Rouge L is best correlated with content verifiability errors (which contains Out of Article Errors) as n-gram matches detect them. Rouge L has negative correlation with semantic frame errors and low correlation with discourse level errors indicating that n-gram matching fails to capture them. We observe that OpenIE is more correlated with semantic frame errors.
The metric matches entities and verifies the predicate that relates them and hence is able to capture semantic frame errors. We observe that BERTScore's high correlation is primarily due to its ability to capture content verifiability errors while it has low correlation on semantic frame errors.

\paragraph{QGA metrics}
Both QGA metrics have negative correlation with discourse errors suggesting that QGA metrics are not able to capture coreference errors or discourse link errors potentially due to the entity oriented questions in their training data. FEQA additionally is also negatively correlated with semantic frame errors and has low positive correlation with content verifiability errors. In contrast QAGS is best correlated with semantic frame errors.
% In both FEQA and QAGS questions are generated around entities so it is not surprising that these errors are not being captured.

\paragraph{Entailment metrics}
Both entailment metrics correlate well with semantic frame and content verifiability errors. DAE has the highest correlation of all metrics with discourse errors suggesting that entailment at the dependency level can help model discourse errors (\CorefE{} and \ConE{}). FactCC is nearly uncorrelated in this category, indicating that artificially generated factual errors need to go beyond simple pronoun swaps to train models to capture discourse errors. FactCC is best at capturing semantic frame among all metrics evaluated.

\section{Related Work}
\label{sec:related_work}
\citet{sumcriticaleval} and \citet{fabbri2020summeval} find that standard n-gram based metrics have low correlation with human judgements of factuality. Motivated by this, several automated metrics falling in two paradigms were proposed to improve the evaluation of factuality.

\paragraph{Entailment Classification}
\citet{goodrich19,factcc,factentailment, goyal2020evaluating} model factuality as entailment classification breaking down the summary into smaller units, such as sentences, which
are verified against the original article. 
However, modeling factuality as a classification task requires supervision on factual and hallucinated data.
FactCC \citep{factcc} is trained on the CNN/DM dataset augmented with four types of artificial mistakes as supervision.

\paragraph{Question Generation and Answering (QGA)}
FEQA \citep{feqa} and QAGS \citep{factqa} are two metrics which reduce factuality evaluation to question generation and answering. These methods use a question generation model to obtain questions from the output summary and a question answering model to answer them, separately using the article and the output summary. 
% In this paradigm, the more answers match the more the output summary is considered to be factual. 

\paragraph{Prior Efforts on Factuality Annotations of Summaries}

\citet{fabbri2020summeval} and \citet{factentailment} have collected annotations on the CNN/DM and XSum dataset respectively. 
In this work we cover both datasets to ensure greater data diversity. 
Other efforts \citep{factcc, factqa, feqa} were smaller in scale \citet{feqa} and \citet{factcc} annotated 200 and 503 sentences while \citet{factqa} annotated 470 summaries (we collect judgements on 2250 summaries).
Crucially, all previous efforts portray factuality as a binary label without variations in degree or type of factual errors.

\section{Conclusion}

In this work we provide a linguistically grounded typology of factual errors which we use to collect \dataset{}, a dataset of human annotations of 2250 summaries covering both CNN/DM and XSum datasets. We use \dataset{} to assess the factuality of summarization systems and benchmark recently proposed factuality metrics highlighting the types of errors they can capture. With the \dataset{} benchmark we have started moving away from a summary-level binary understanding of factuality.

%\balance

\section{Ethical Considerations}
We have collected crowd annotations using the Amazon Mechanical Turk platform. Workers were paid 50\% more than the average American minimum wage and offered additional bonuses as an incentive to maintain high quality work. No information about the workers will be released and worker IDs will be anonymized.

\section*{Acknowledgements}
The authors are grateful to the anonymous reviewers for their feedback, and to Anjalie Field, Rishabh Joshi, Alissa Ostapenko, Dheeraj Rajagopal, Evangelia Spiliopoulou, Shuly Wintner, and the members of the Tsvetshop group for their invaluable feedback and support in various stages of the project.
This material is based upon work supported by the DARPA CMO under Contract No.~HR001120C0124, and in part by the National Science Foundation under Grants No.~IIS2040926 and No.~IIS2007960.  Any opinions, findings and conclusions or recommendations expressed in this material are those of the author(s) and do not necessarily state or reflect those of the United States Government or any agency thereof.
\bibliography{anthology,references}
\bibliographystyle{acl_natbib}

\cleardoublepage

\appendix

\section{Appendices}
\label{sec:appendix}
\subsection{Model details}
\label{sec:model-details}
We provide details of the models used in the human evaluation task to construct \dataset{}.
\subsubsection{CNN/DM datset}
On the CNN/DM \citep{cnn-dm} dataset we use five different models. We use the preprocessed model outputs provided by \citet{fabbri2020summeval}.
\textbf{S2S} an LSTM based Sequence-to-Sequence with attention model \citep{rush}\\
\textbf{PGN} an LSTM based Pointer-Generator Network with Copy Mechanism \citep{pgn}\\
\textbf{BUS} Bottom-Up Summarization \citep{bus} - a Pointer-Generator model with a data-efficient content selector to over-determine phrases in a source document that should be part of the summary.\\
\textbf{BERTSum} summarization with pretrained encoders \citep{liu-lapata-2019-text}\\
\textbf{BART} \citep{bart}\\

\subsubsection{XSum dataset}
On the XSum dataset \citep{xsum} we use four different models. All model outputs for this dataset are taken from \citep{factentailment}
\textbf{PGN} pointer-generator network from above \citep{pgn}\\
\textbf{TConvS2S} Topic-Aware Convolution Sequence-to-Sequence \citep{xsum} \\
\textbf{TranS2S} A randomly initialized Transformer \citep{vaswani2017attention} encoder-decoder model fine-tuned on the XSum dataset \\
\textbf{BERTS2S} Transformer encoder-decoder model with parameter sharing \citep{rothe-etal-2020-leveraging} where both encoder and decoder are initialized with the BERT-Base checkpoints \citep{bert} and fine-tuned on XSum\\

\subsection{Metrics}
\label{sec:metrics_appendix}
In this work we compare the following five metrics.

\paragraph{BERTScore \citep{bertscore}: }
We report BERTScore Precision, Recall, and F1 between the model output and the article being summarized. Our experiments show that recall and F1 do not correlate as well with the human judgement of factuality for BERTScore.  

\paragraph{OpenIE}:  We use a simple baseline based on OpenIE \citep{openie} and Sentence-BERT \citep{sentence-bert}. We use OpenIE \citep{openie} to extract subject-relation-object triplets from the article, reference summary, and model generated summary. We consider binary relations only and thus use the first two arguments of the relation.\footnote{We use the model and implementation from \cite{stanovsky-etal-2018-supervised} for OpenIE extraction.} After replacing corefering entity mentions with the main mention of the cluster\footnote{\url{https://github.com/huggingface/neuralcoref}}, we use BERT base Sentence-BERT \citep{sentence-bert} to obtain embeddings of each element of the subject-relation-object triplets extracted by OpenIE. Two relation triplets are considered to be equivalent if their embeddings have cosine similarity higher than a threshold for all three elements of the triplet (we use 0.6 as threshold after a grid search between 0.5 and 0.9 on data from our pilot study).

\paragraph{FEQA \cite{feqa}: } FEQA is a question generation and answering (QGA) factuality metric. We relied on the original implementation of the authors for this metric as well as their pre-trained model weights. We used the full summary to generate questions and we answer them both using the summary and article text.

\paragraph{QAGS \cite{factqa}: } QAGS is another QGA metric. The authors kindly provided outputs on the \dataset{} benchmark generating 10 questions for each summary.

\paragraph{DAE \cite{goyal2020evaluating}: } DAE is an entailment classification metric that operates on dependencies. The authors kindly provided outputs on the \dataset{} benchmark. We note that the model was trained with a max length of 128 after concatenating both article and summary. The CNN/DM articles can be significantly longer, thus the results reported for this metric involve truncating parts of the article.

\paragraph{FactCC \cite{factcc}: } FactCC is an entailment classification metric. We use the sentences of the model generated summary as input claims to the entailment classifier FactCC. For each sentence we obatain a binary factuality label. We take the average of these labels as the factuality score for the summary.

%\paragraph{OpenIE Baseline:} \todo{I might skip OpenIE baseline} We also propose a simpler version of our method based on OpenIE \citep{openie} and Sentence-BERT \citep{sentence-bert}. This baseline helps to isolate the benefits of learning a contextual representation of relations instead of using existing sentence embedding models. We use OpenIE \citep{openie} to extract subject-relation-object triplets from the article, reference summary, and model generated summary. Since we consider binary relations, we only use the first two arguments of the relation.\footnote{We use the model and implementation from \cite{stanovsky-etal-2018-supervised} for OpenIE extraction.}
%After replacing corefering entity mentions with the main mention of the cluster\footnote{\url{https://github.com/huggingface/neuralcoref}}, we use BERT base Sentence-BERT \citep{sentence-bert} to obtain embeddings of each element of the subject-relation-object triplets extracted by OpenIE. Two relation triplets are considered to be equivalent if their embeddings have cosine similarity higher than a threshold for all three elements of the triplet. This essentially entails matching the two entities and verifying that the relation expressed by the predicate of each triplet is semantically similar through sentence embedding. 
%We manually tune the threshold on a validation set of 150 annotations (50 articles and 3 different models annotated by a single annotator). \todo{Value of threshold} 

\subsection{Summarization System Analysis Details}
See \autoref{tab:framework} for more details.
\begin{table*}
\centering
%\resizebox{\linewidth}{!}{
\begin{tabular}{l | c | c c c c c c c c}
\toprule
% & \multicolumn{7}{c}{Factuality} \\
& Incorrect & PredE & EntE & CircE & CorefE & LinkE & OutE & GramE & Other \\
\hline
Seq2Seq     & 74.8\% & 11\% & 46\% & 13\% & 15\% & 5\% & 14\% & 24\% & 0\%\\
PGN         & 26.5\% & 4\% & 46\% & 0\% & 39\% & 0\% & 4\% & 21\% & 0\%\\
Bottom Up   & 62.6\% & 6\% & 56\% & 6\% & 17\% & 9\% & 6\% & 21\% & 4\%\\
BERTSum     & 27.2\% & 10\% & 37\% & 10\% & 23\% & 3\% & 13\% & 10\% & 0\%\\
BART        & 23.8\% & 4\% & 25\% & 8\% & 33\% & 4\% & 17\% & 17\% & 4\%\\
\hline
PGN         & 96.9\% & 16\% & 28\% & 11\% & 1\% & 1\% & 34\% & 13\% & 0\%\\
TConvS2s    & 89.8\% & 10\% & 24\% & 18\% & 1\% & 0\% & 45\% & 1\% & 0\%\\
TranS2S     & 96.9\% & 10\% & 32\% & 15\% & 0\% & 0\% & 44\% & 1\% & 0\%\\
BERTS2S     & 83.7\% & 10\% & 25\% & 23\% & 0\% & 0\% & 38\% & 3\% & 0\%\\
\hline
All models  & 60.0\% & 10\% & 36\% & 13\% & 10\% & 3\% & 27\% & 12\% & 1\%\\
\bottomrule
\end{tabular}
%}
\caption{Proportion of summaries that include factual errors, with breakdown of the categories of errors according to our human study. F8 corresponds to errors that are not captured by our typology. Full specification of categories of errors in \autoref{tab:framework}.}
\label{tab:factcategory}
\end{table*}
%}

\subsection{Hotelling Williams Test}
The correlation numbers in \autoref{tab:results} should be read in combination with the pairwise Hotelling-Williams test \citet{graham-2015-evaluating} results in \autoref{tab:metric-metric-results}. The highlighted numbers indicate pairs of models for which the difference in correlation is statistically significant. We use partial correlations to run the test and compute metric-metric correlations.

\begin{table*}
\centering
% \resizebox{\linewidth}{!}{
\begin{tabular}{l| c c c c c c c c c c c c} %c }
\toprule
& B & MET& R-1 & R-L & BS-P & OpIE & FEQA & QAGS & DAE & FCC \\
\midrule
BLEU & - & \textcolor{green}{0.82} & \textcolor{green}{0.77} & \textcolor{green}{0.85} & \textcolor{green}{0.12} & 0.25 & \textcolor{green}{0.03} & -0.02 & \textcolor{green}{0.05} & \textcolor{green}{0.06} \\
METEOR & \textcolor{green}{0.82} & - & 0.87 & 0.85 & \textcolor{green}{0.17} & 0.27 & \textcolor{green}{0.02} & \textcolor{green}{-0.02} & 0.09 & \textcolor{green}{0.07}\\
Rouge-1 & \textcolor{green}{0.77} & 0.87 & - & 0.89 & \textcolor{green}{0.22} & 0.21 & \textcolor{green}{0.01} & \textcolor{green}{-0.03} & 0.09 & \textcolor{green}{0.07}\\
Rouge-L & \textcolor{green}{0.85} & 0.85 & 0.89 & - & \textcolor{green}{0.18} & 0.21 & \textcolor{green}{0.01} & \textcolor{green}{-0.04} & 0.08 & \textcolor{green}{0.07} \\
BERTS P & \textcolor{green}{0.12} & \textcolor{green}{0.17} & \textcolor{green}{0.22} & \textcolor{green}{0.18} & - & \textcolor{green}{0.20} & \textcolor{green}{0.01} & \textcolor{green}{0.06} & \textcolor{green}{0.18} & \textcolor{green}{0.27}\\
OpenIE & 0.25 & 0.27 & 0.21 & 0.21 & \textcolor{green}{0.20} & - & \textcolor{green}{-0.01} & \textcolor{green}{0.09} & \textcolor{green}{0.10} & \textcolor{green}{0.15}\\
FEQA & \textcolor{green}{0.03} & \textcolor{green}{0.02} & \textcolor{green}{0.01} & \textcolor{green}{0.01} & \textcolor{green}{0.01} & \textcolor{green}{-0.01} & - & \textcolor{green}{-0.01} & \textcolor{green}{0.03} & \textcolor{green}{0.04} \\
QAGS & -0.02 & \textcolor{green}{-0.02} & \textcolor{green}{-0.03} & \textcolor{green}{-0.04} & \textcolor{green}{0.06} & \textcolor{green}{0.09} & \textcolor{green}{-0.01} & - & \textcolor{green}{0.07} & \textcolor{green}{0.10} \\
DAE & \textcolor{green}{0.05} & 0.09 & 0.09 & 0.08 & \textcolor{green}{0.18} & \textcolor{green}{0.10} & \textcolor{green}{0.03} & \textcolor{green}{0.07} & - & 0.10\\
FactCC & \textcolor{green}{0.06} & \textcolor{green}{0.07} & \textcolor{green}{0.07} & \textcolor{green}{0.07} & \textcolor{green}{0.27} & \textcolor{green}{0.15} & \textcolor{green}{0.04} & \textcolor{green}{0.10} & 0.10 & -\\
\bottomrule
\end{tabular}
% }
\caption{Pearson correlation between metrics. If value is in green, the metrics are not the same significant to the 0.05 threshold with the Hotelling Williams test.}
\label{tab:metric-metric-results}
\end{table*}

\subsection{Mutual Exclusiveness of typology:} 
\begin{figure}
    \centering
    \includegraphics[width=\columnwidth]{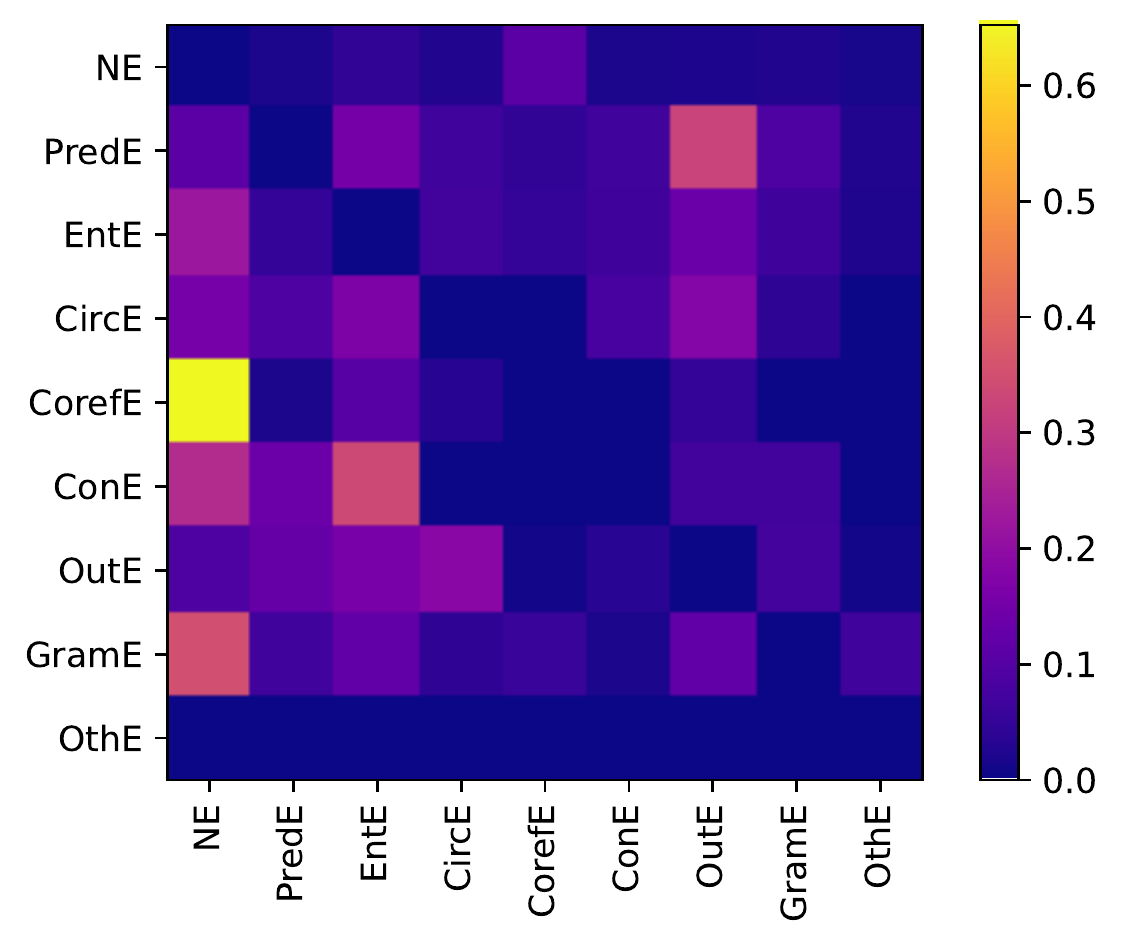}
    \caption{Confusion matrix of different types of errors. Entry at row $i$, column $j$ corresponds to the frequency of annotations that have F$i$ as the majority class and for which disagreeing annotator selected F$j$.}
    \label{fig:confusion}
\end{figure}

To understand if our annotations are mutually exclusive, we study cases where two annotators agree on the error category (majority class) and one disagrees (minority class). In \autoref{fig:confusion}, we report the confusion between majority and minority classes. For each category as majority, we report the distribution of other categories as minority.

% for annotations with given majority class, we report the frequency with which disagreeing annotators selected another class. In other words, the frequency with which the majority class was confused with other classes. We omit the cases where all three annotators selected different categories since we have no means of determining which one was correct.

% F0
We observe that all categories with the exception of \textbf{\OutE{}} are frequently confused with \textbf{\NE} which stands for no factual error. This primarily due to the noise in the annotations collected by crowd workers. However, for category \textbf{\CorefE{}} (coreference errors) the confusion is significantly higher with 69.7\%. We have noticed the same trend in practice tutorials: crowd annotators easily overlook situations where the correct pronoun is used (in terms of number and gender) but no antecedent appears in the summary. Intuitively after reading the article, unless paying particular attention, it is easy to subconsciously associate referring expressions with entities in the article without noticing their absence in the summary. The error persists despite stating the scenario explicitly in the instructions. This indicates an issue with annotators rather than annotation scheme.

% F4
The other trend that we observe is that categories \textbf{\PredE{}} (wrong relation) and \textbf{\CircE{}} (wrong modifier) are often confused with \textbf{\OutE{}} (outside information). In our definition of \textbf{\OutE{}}, outside information corresponds to the presence of entities not mentioned in the article or relations that cannot be verified based on the article. The confusion with \textbf{\PredE{}} indicates that annotators can have different judgements on whether a relation is verifiable based on the article. Similarly, but to a lesser degree, wrong circumstantial information might be considered unverifiable given the article.

% F7
Finally, there were relatively few  discourse context errors \textbf{\ConE{}}, so the analysis is less statistically significant. Discourse context errors correspond to using a wrong connectors between different facts, for example different logical links. These were confused with \textbf{\PredE{}} and \textbf{\EntE{}} (wrong relation). The distinction between the two errors lies in the confusion between what an entity and a fact are, since \textbf{\PredE{}} occurs at the frame level while \textbf{\ConE{}} at the discourse level. Note, that there was no confusion in the other direction (\textbf{\PredE{}} being confused with \textbf{\ConE{}}).
 
\subsection{Annotation Setup Details}
\label{sec:annotation_details}
Below are more details on the annotation set up.
\paragraph{Clear Instructions}
We explain the annotation scheme without assuming linguistic knowledge and give several examples for each category. We also provide a practival ste-by-step to determine the category of the errors.

\paragraph{Training}
Every first-time user has to go through a tutorial which exercises the comprehension of the annotation scheme. The tutorial presents an article and several hand-crafted summaries of the article that need to be annotated. It is designed to be very similar to the actual annotation task and to contain at least one occurrence of each category of error. Feedback is provided when a user selects the wrong category of error. This tutorial is not used to evaluate users, only to help them understand the different categories in a practical setting.

\paragraph{Qualification test}
To participate in the annotation, users have to obtain a minimum score of 85\% on a qualification test. The test comprehends an article and several summaries to be annotated. It contains at least one instance of each category of error.  
We use this test to verify that users can effectively recognize error categories.
This ensures that users are able to perform the task correctly, but does not enforce that high standards of work quality are maintained throughout the annotation task.

\paragraph{Continuous evaluation}
We continuously evaluate a user by verifying that they read the text. For every article that is annotated, we ask to identify one of three entities that was not present in the article. We also monitor the annotations on artificially altered sentences that are randomly inserted at the end of summaries. Wrong sentences contain one of the following errors: negation of declarative sentences (\PredE), pronoun swap (\CorefE), sample sentence from another article (\OutE), word scrambling (\GramE). We immediately block users that fail the entity test or perform poorly on these sentences (less than 50\% of correct answers on altered sentences) to ensure high quality annotations.

\paragraph{Bonuses}
All workers are paid 50\% more than the average American minimum wage but we offer bonuses for scores of 60\% or above on the continuous evaluation, and for completion a sequences of 10 annotations. We observe that bonuses increase the percentage of users with high continuous evaluation scores ($<$10\% blocked users with bonuses versus 30\% without bonuses).

\subsection{Correlation with Confounding Variables} 
Partial correlation measures the degree of association between two random variables, with the effect of a set of controlling random variables removed. Although we are unaware of the exact confounding variable, we use the categorical variable $C$ of which system and dataset the summary was generated from.  
%We use the categorical variable of which model and dataset a summary was generated from as control variable.

Let $M_k$ represent the output of metric $k$ on the summaries. To compute partial correlation between $M_k$ and human judgements $H$ which we treat as random variables, we solve the two regression problems $M_k | C=c \sim w_{M_k} c$ and $H | C=c \sim w_{H} c$ and get the residuals:
$$\Delta{M_k} = M_k - \hat{w}_{M_k} C$$
$$\Delta{H} = M_k - \hat{w}_{H} C$$
And then calculate the correlation between these residuals $\rho(\Delta{M_k}, \Delta{H})$ instead of the original random variables. Since partial correlations are proper correlations between random variables, we can apply statistical significance tests without any modification.

\subsection{Annotation Interface}
We include screenshots of the annotation interface which we will make available. 

\begin{figure*}
    \centering
    \includegraphics[width=\linewidth]{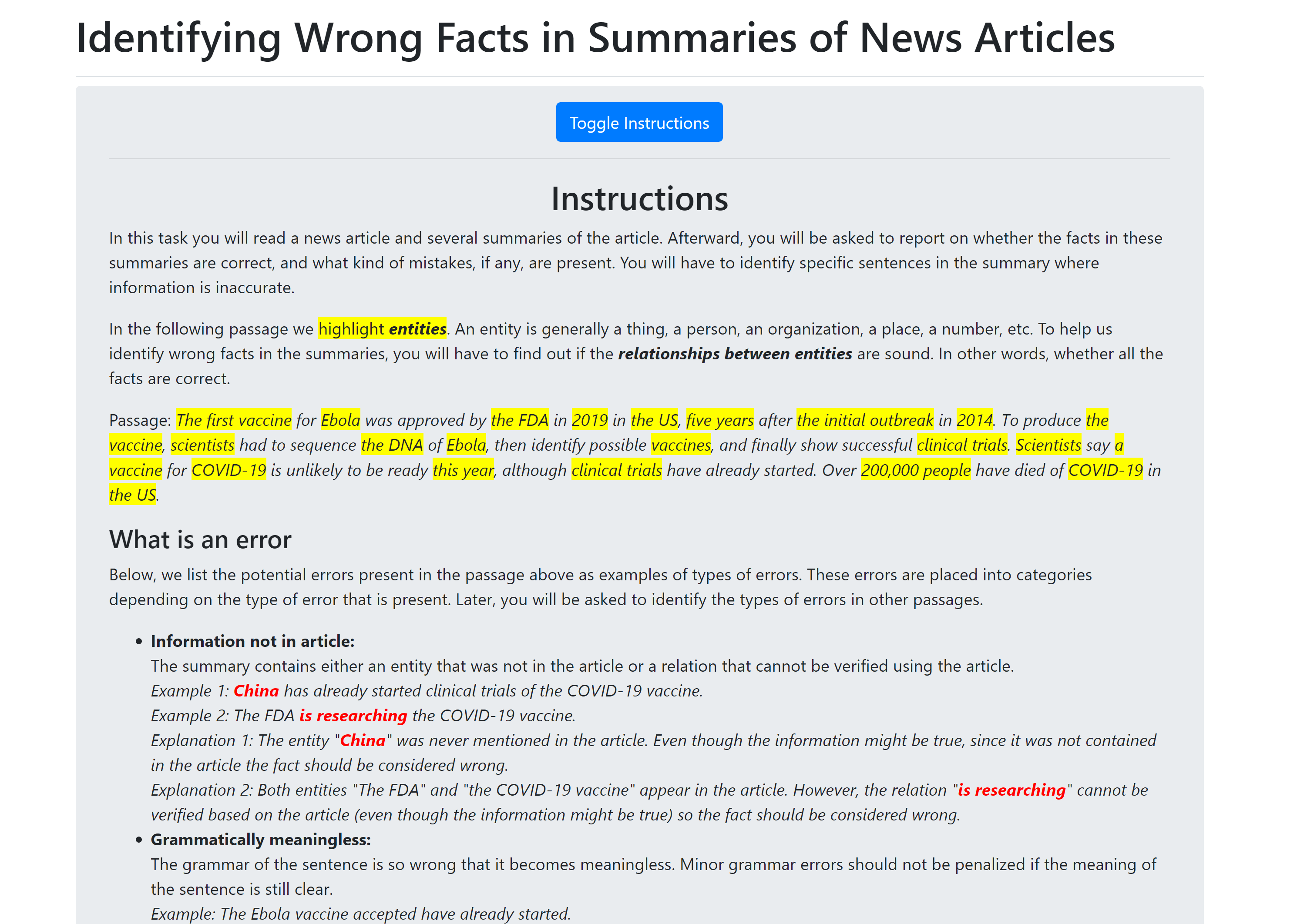}
    \caption{Instructions can be toggled.}
\end{figure*}
    
\begin{figure*}
    \centering
    \includegraphics[width=\linewidth]{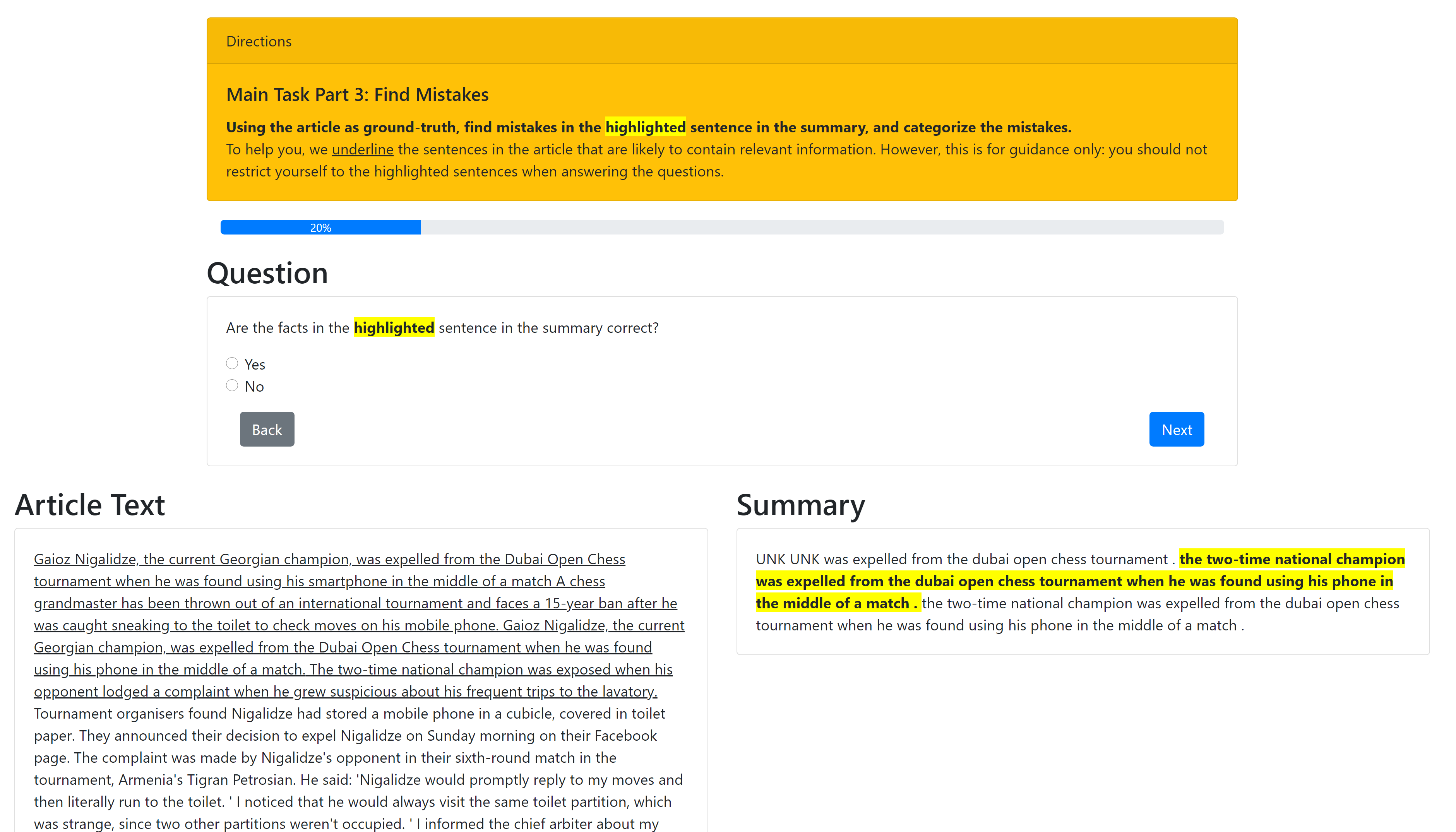}
    \caption{The sentences being annotated is highlighted in yellow. Relevant text is underlined in the article plain text.}
    \label{fig:interface}
\end{figure*}

\begin{figure*}
    \centering
    \includegraphics[width=\linewidth]{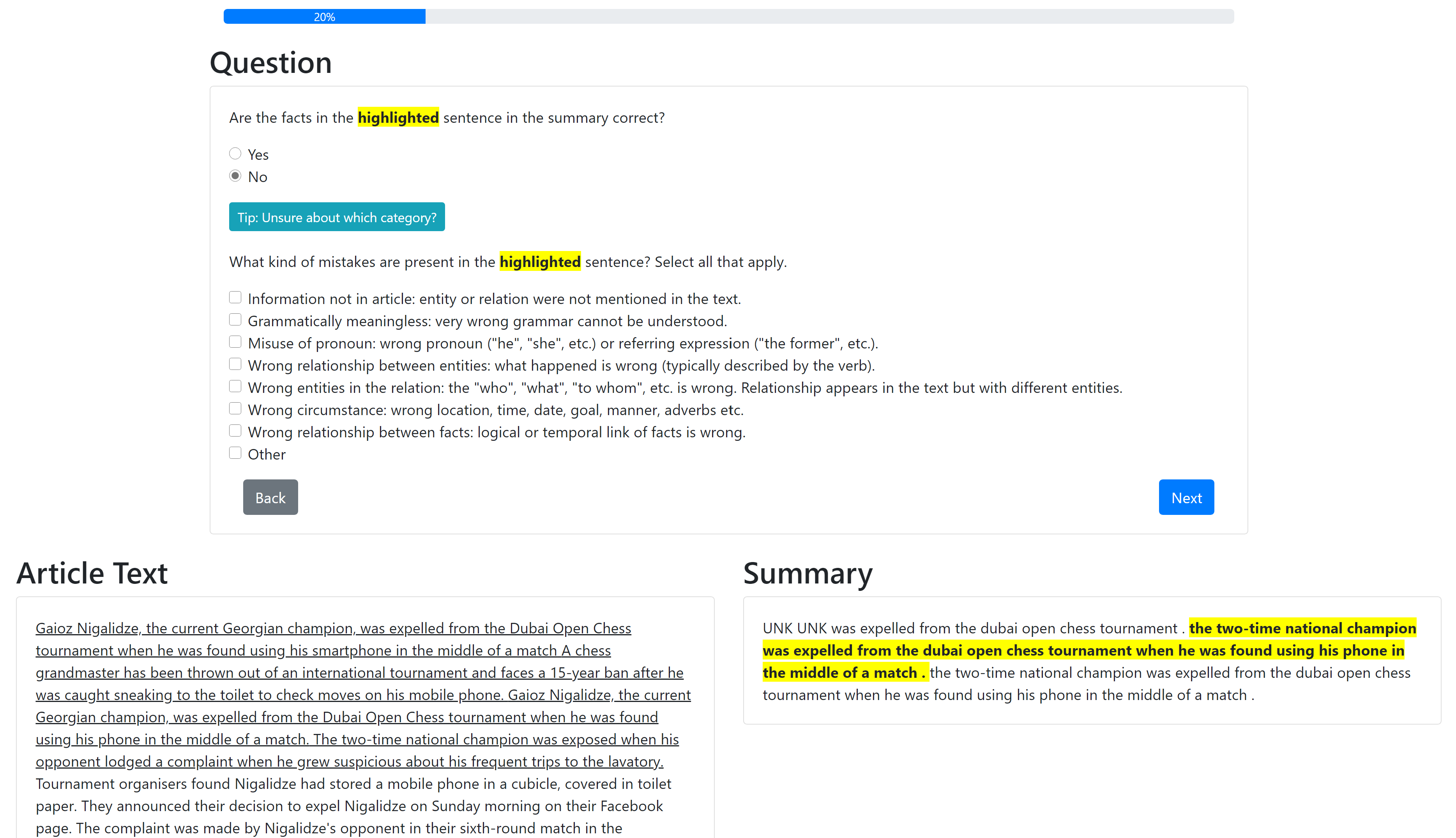}
    \caption{After selecting that the sentence is not factual annotators choose the category of error.}
    \label{fig:interface1}
\end{figure*}
\begin{figure*}
    \centering
    \includegraphics[width=\linewidth]{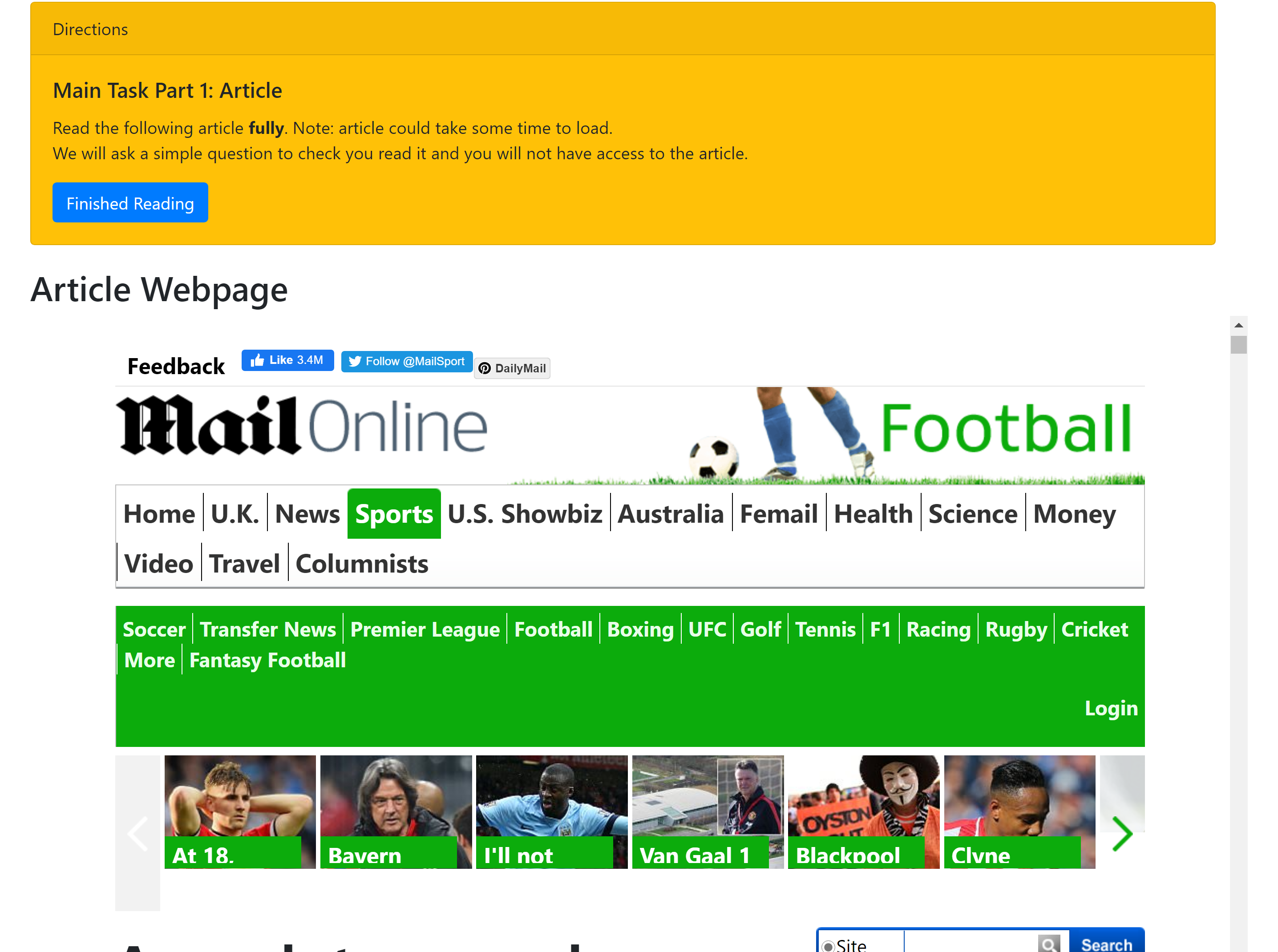}
    \caption{Articles web pages are provided.}\end{figure*}

\begin{figure*}
    \centering
    \includegraphics[width=\linewidth]{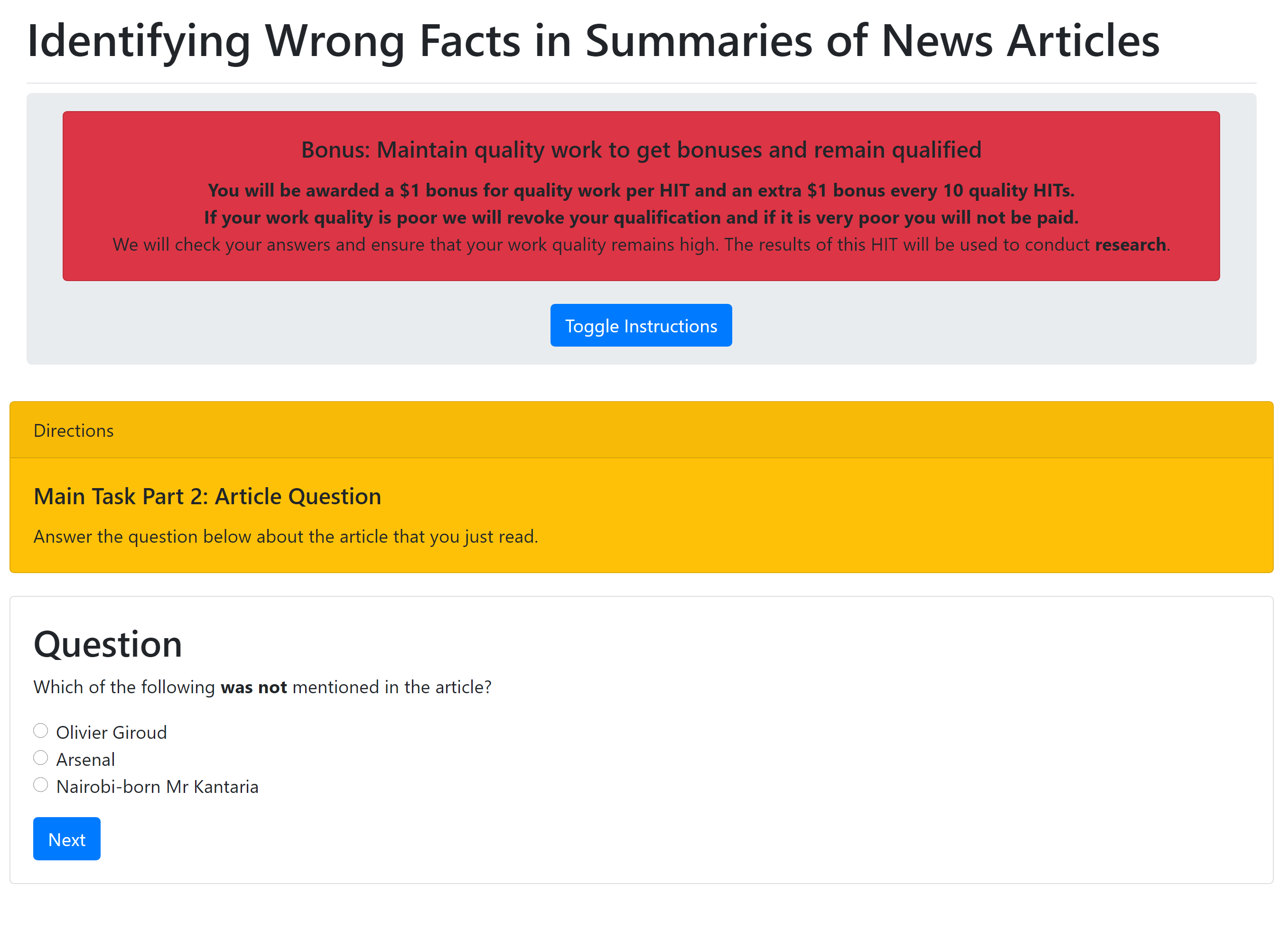}
    \caption{Entity question to ensure annotators read the text.}
\end{figure*}
\end{document}